# Another Look at Quantum Neural Computing

Subhash Kak

## ABSTRACT

The term *quantum neural computing* indicates a unity in the functioning of the brain. It assumes that the neural structures perform classical processing and that the virtual particles associated with the dynamical states of the structures define the underlying quantum state. We revisit the concept and also summarize new arguments related to the learning modes of the brain in response to sensory input that may be aggregated in three types: associative, reorganizational, and quantum. The associative and reorganizational types are quite apparent based on experimental findings; it is much harder to establish that the brain as an entity exhibits quantum properties. We argue that the reorganizational behavior of the brain may be viewed as inner adjustment corresponding to its quantum behavior at the system level. Not only neural structures but their higher abstractions also may be seen as whole entities. We consider the dualities associated with the behavior of the brain and how these dualities are bridged.

## INTRODUCTION

I proposed the term *quantum neural computing* [1] in 1995 to suggest that the complete neural system of an organism defines a *whole* and it can be viewed as a quantum system at its deepest level and a classical system at the embodied level. Whereas its neural connections constitute the conscious system which is classical, the quantum system is supported by the virtual particles associated with its dynamic states. The philosophical idea behind this proposal was that like material objects abstract entities also have a reality that is subject to quantum laws. This proposal was not to suggest that the brain's structures supported quantum coherent processes in the network of neurons but rather that the brain's functioning as a whole was according to quantum laws at a deeper level.

Soon afterwards Karl Pribram invited me to participate in a conference he had organized at Radford University on Learning as Self-Organization [2] for which he had brought together leading neuroscientists and where my role as an outsider to that group was to give a broad systems perspective on the problem. Believing that the idea of quantum neural computing could be reiterated in the setting of learning in the brain, I chose the title *The three languages of the brain: quantum, reorganizational, and associative* for my presentation [3] in which I argued that in addition to self-organization, one must also consider associational and quantum aspects of learning. The choice of the number three for languages was to aggregate complex behavior of the brain into three main classes and not meant as some fundamentally established number.

In the mainstream view, learning in the brain is taken to be based on the Hebbian mechanism. But the operation of this mechanism remains a puzzle. I proposed that the Hebbian mechanism emerges from the adjustments of the brain as a quantum system to alter conditions within that make its behavior consistent in an inside-the-box view.

That the structure of the brain reflects associative and reorganizational learning is quite apparent from experimental findings; it is much harder to establish conclusively by considering physiological evidence that the brain as a whole exhibits quantum properties. For example, the model of quantum activity based in microtubules originally proposed by Hameroff [4] with the further notion of reduction by gravity by Penrose [5] has not found favor with researchers (e.g. [6]). The idea of reduction of microtubular quantum states does not change the character of the brain from that of a machine and, therefore, it would remain subject to the shortcomings that such a model has from the point of view of agency and awareness.

The proposition that one of the languages of the brain is quantum is not to imply that the brain is a quantum computer with data stored as superpositions of 0s and 1s. Quantum theory as the deepest theory of reality should apply to all systems, large and small, although for a large system its interaction with the environment and within its own subparts would destroy quantum coherence. If the traces on the screen in the double-slit experiment are the representation of the quantum reality on the two-dimensional surface of the screen, the structure of the brain may be seen as a representation of the underlying quantum reality on the three-dimensional system of neurons. Like additional incoming photons alter the pattern on the screen, new sensory impressions change the pattern of connections amongst the neurons.

We are arguing for what may be termed "inside-the-box view" of the problem of brain behavior. To appreciate this, imagine the problem of the dynamics of the constituent quarks and gluons within the proton as it interacts with its environment. We propose that although the behavior of the proton represented as the wave function might suffer jumps or discontinuities, when viewed from within the proton (although such a viewing may not be possible with the science and technology available to us) the evolution has no such discontinuities. For example, the proton may be taken to travel over both paths in a Mach-Zehnder interferometer, but from within the proton, there would be no such division and its behavior would remain consistent. In the inside-the-box view, the structure of the proton will experience appropriate adjustments that are consistent within the framework of the interactions within.

I have advanced the "inside-the-box view" in different forms for several years [8]-[12], and articulated it both in the context of particles as well as neural systems. I have suggested that neural structure itself is a consequence of the underlying quantum reality and that attempts to determine it completely change it. The system and the environment evolve in an ecological manner. Just as the electron orbit states of the hydrogen atom are fixed by quantum theory even if they should not be occupied, the evolution of the system exhibits discrete jumps fixed by the underlying law.

According to this view, the reductionist and the system-as-a-whole position are complementary and one cannot be privileged over the other. If the properties of the components go into defining the properties of the system (Figure 1), the integral properties of the system equally affect the properties of the components.

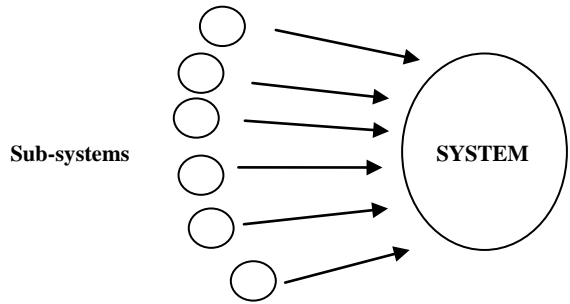

**Figure 1**. System viewed as a sum of its parts

The new system level behavior is not to be seen as emerging from the collection of its parts, but rather as a characteristic of the system as a unity. In this understanding, a computer no matter how complex will never be able to have the same behavior as the brain because neither the size nor the complexity of interconnections are to be seen as leading to the specificity of brain behavior although they may be pre-requisites. Brain behavior is to be seen as a unity at a higher level and, equivalently, as interactions between its various sub-parts.

The properties of the components can only be determined in relation to the system as a whole. Reducing the system to its parts (Figure 2) loses essential information, which is compensated for by the postulation of new and unexpected emergent properties.

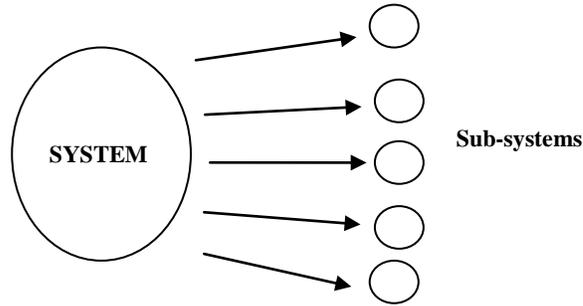

**Figure 2**. System reduced to its sub-systems

Shaffer [7] has argued for a nonempirical domain of physical reality that is a realm of nonmaterial forms. He claims that this realm is the realm of potentiality in physical reality as an indivisible wholeness, and it is the ultimate reality because everything empirical is the actualization of its forms. He further suggests that the brain is sensitive to the potentiality waves in this realm. He acknowledges that this scheme is speculative since there is no experimental evidence in support of this nonmaterial realm.

Although Shaffer's ideas have some overlap with our view, we are not seeking to go beyond the inferred physical structure of the system under consideration and, therefore, we do not consider ontological questions regarding the ultimate nature of reality.

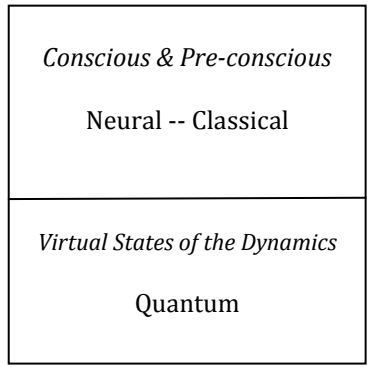

**Figure 3**. Brain viewed as a dual quantum-classical system

One can determine whether a "small" system is operating according to quantum laws by experiments related to superposition of the relevant states and the obtaining of outcomes in a "probabilistic" fashion upon interaction with the system. The situation with a "large" system, whose structure is accessible to experimentation, is more troublesome. In small systems such as a proton, the internal structure in terms of quarks is not taken into consideration in accounting its behavior as a quantum object, but there is no reason why one should not ask as to what happens to the internal structure in an interaction. The structure of the large system is accessible in principle, and one needs a consistent narrative to explain the specific output.

The behavior of the brain has several modes mirroring the multiplicity of the modes of a living system [12],[13]. A non-living system, no matter how complex, can, in principle, be described fully. A living system needs several *languages* because its behavior is active and it is associated with attention and agency. This sets up a duality since selectively concentrating on a part of the environment implies ignoring of other parts.

It appears that a generalization of the complementary principle is needed to deal with the behavior of *large* quantum systems. It may be recalled that Niels Bohr used the concept of complementarity [14],[15] to imbed quantum mechanics in a coherent and rational framework. Since the final representation of observations is in our sense experience that is described in classical terms, there is a logical incompatibility between the quantum process and its observation. According to complementarity, a process may be explained as a particle picture or a wave picture, but not both, simultaneously.

Bohr suggested that "independent reality in the ordinary physical sense can neither be ascribed to the phenomena nor to the agencies of observation. After all, the concept of observation is in so far arbitrary as it depends upon which objects are included in the system to be observed. Ultimately, every observation can, of course, be reduced to our sense perceptions." ([14], p. 53) He elaborated this in a later essay: "Within the scope of classical physics, all characteristic properties of a given object can in principle be ascertained by a single experimental arrangement, although in practice various arrangements are often convenient for the study of different aspects of the phenomena. In fact, data obtained in such a way simply supplement each other and can be combined into a consistent picture of the behaviour of the object under investigation. In quantum mechanics, however, evidence about atomic objects obtained by different experimental arrangements exhibits a novel kind of complementary relationship. Indeed, it must be recognized that such evidence which appears contradictory when combination into a single picture is attempted, exhaust all conceivable knowledge about the object. Far from restricting our efforts to put questions to nature in the form of experiments, the notion of *complementarity* simply characterizes the answers we can receive by such inquiry, whenever the interaction between the measuring instruments and the objects form an integral part of the phenomena." ([15], p.4)

Quantum mechanics is not a representation of reality, but rather of our knowledge of this reality. With this as background, we first note that biological systems do exhibit individuality, and, therefore, their behavior may be seen as a consequence of effort, will, and intention [16],[17],[18] or a learnt response [19], which are complementary pictures. Neither of these views is sufficient to explain all behavior [20]-[23] and thus the situation is analogous to that of duality of quantum theory. A large quantum system may be seen, by an extension of the idea of complementarity, either as responding in a random fashion to a stimulus when seen as a unity, or as reorganizing itself so that the behavior can be accounted for in the reorganized structure. Representation of brain behavior may be done in terms of self-organization or equivalently as a consequence of the agency of an individual in the inner theater of the brain.

Looking at the matter of adjustment within the neural system, I shall show that the retrieval of memories from single neurons is possible even though memories are distributed in the synaptic interconnection strengths amongst the neurons.

The inside-the-box view of brain function, although described qualitatively, provides valuable insights into behavior. It helps us see the dualities of behavior and how these dualities are bridged by abstraction at a higher level. At their most fundamental level, these dualities emerge from the dual conception of the brain (Figure 3) which is a classical system with a quantum ground. This structure is mirrored in the dual conception of the conscious and the unconscious at one level and several other polarities associated with mind and behavior.

## LEARNING THROUGH ASSOCIATION AND SELF-ORGANIZATION

The Hebbian model of neural network learning [24],[25] assumes a structure consisting of neurons for which the synaptic interconnections are strengthened or weakened during training. This is at the basis of several neural network training algorithms that have found applications in signal processing. Conversely, models of neural network memory have feedback connections trained by Hebbian learning without consideration to physical separation amongst the neurons and the delay caused by the propagation of spike activity.

Although associative learning itself may be viewed as a special case of self-organization, we imply by it the modifications of the synaptic interconnection strengths within the modular structures of the brain. By self-organization proper we mean the evolution of the modules in relation to other modules.

There exists a dichotomy between distributed and localized memories, each of which appears to be of different type. Memories also have a broad division into short term and long term [26]. Many artificial neural network learning models are implicitly long term memories in the sense that they take a considerable time to train [25],[27]-[33]. There is of course another network that does provide instantaneous learning [23],[31] that may be seen as being analogous to short term memory.

In popular models of memory, the neural network is trained by Hebbian learning, in which the connectivity (synaptic strength) of neurons that fire together strengthens and that of those that don't gets weakened. The interconnection matrix $T = \sum x^{(i)} x^{(i)t}$, where the memories are column vectors, $x^i$, and the diagonal terms are taken to be zero.

A memory is stored if

$$x^i = \text{sgn}(Tx^i) \tag{1}$$

where the *sgn* function is 1 if the input is equal or greater than zero and -1, if the input is less than 0. We know from computer experiments on this model that the memory capacity of such a network is about 0.15 times the number of neurons. In addition to the stored desired memories, their complements will be stored (although not all because of the asymmetry of the sgn operation), as also some spurious memories [22],[24],[25].

In the generator model [25],[33], the memories are recalled by the use of the lower triangular matrix $B$, where $T = B + B^t$. Effectively, the activity starts from one single neuron and then spreads to additional neurons as determined by $B$ (Figure 4). With each update, the fragment enlarges by one neuron and it is fed back into the circuit. The spreading function where the information spreads from neuron 1 to neuron 2 and so on has an implicit assumption regarding the geometrical (proximity) relationship amongst the neurons. In a variant of the above, one may assume that the activity does not start with a single neuron, but with a collection of neurons.

The physical organization of any neural network makes the separations amongst the specific neurons to have many different values. The activity would have a tendency to flow in different directions for different starting points. It should be recognized that the neurons separations need not satisfy Cartesian constraints since neural pathways may be coiled up or twisted in a three-dimensional geometry. The proximity matrix $P$ represents the distances between all pairs of neurons.

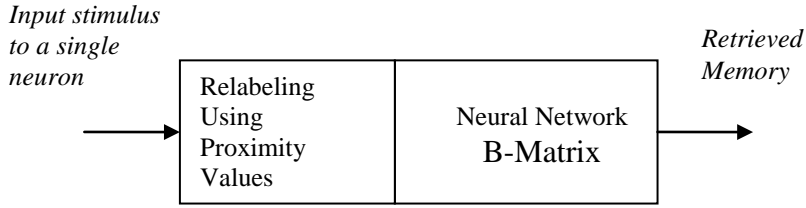

**Figure 4**. Use of proximity values to determine the neural memory

For the spreading of activity from any specific neurons, one must first compute the proximity values to the other neurons to determine how the activity will spread.

Starting with the fragment $f^i$, the updating proceeds as:

$$f^i \ (new) = \text{sgn} \ (Bf^i \ (old) \ ) \tag{2}$$

in which the original fragment values are *left unchanged* on the neurons and the updating proceeds *one step at a time*. Thus the activity will expand from one to two neurons and then on to three neurons, and so on.

It is reasonable to assume that in a network of densely connected neurons, activity originates in some area and spreads to others. Neurons belonging to different physiological structures are associated with different functions and their organization is characterized by category and hierarchy. As to neuron activity, after firing, they enter a phase of refractoriness.

Feedforward networks are naturally associated with spreading activity because the direction of activity is specified *a priori*. The B-matrix model extends the notion of spreading of activity to feedback networks. The order in which the activity will spread depends on the propagation time for the spiking activity to reach neighboring neurons, which is reflected in the proximity values associated with the network.

## QUANTUM SYSTEM REPRESENTED AS A SELF-ORGANIZING SYSTEM

Here we consider the question of how a large quantum system may be represented internally by a classical system. Consider, for simplicity, the wave function $| \Psi \rangle = \sum_{1}^{n} a_i \ | \Psi_i \rangle$ . Suppose the measurement leads to a binary variable, then the reduction will be viewed as a process that led to the reorganization of the system into a total of *n* different cases. The collapse into the *n* basis states may be seen, equivalently, as a neural system that is associated with *n* different outputs.

Note that this view is different from that of Perus and others [34] who consider an associative memory obtained in analogy with the reduction of the wave function.

More specifically, consider the qubit $(a \ | 0 \rangle + b \ | 1 \rangle)$, which upon interaction with the measurement device will collapse to $| 0 \rangle$ or $| 1 \rangle$ .

Since an infinity of values of *a* and *b* in the input are mapped into just the two outputs $| 0 \rangle$ or $| 1 \rangle$ , it would be assumed that the system reorganized itself so as to provide the correct output. If *a* and *b* take *n* values each, the total number of reorganizations that need to be considered to account for the variety of outputs is

$n^2$

that is obtained as the number of total cases $2 \times n^2$ divided by 2 to account for the mutual probability amplitude constraint amongst $a$ and $b$.

The collapse of the function for a specific output could be seen as the emergence of a resonance within the system.

## THE QUANTUM GROUND

Different theories have been proposed to describe the nature of the quantum ground. Fröhlich [35] showed how there could be long range coherence in biological systems. In 1967, Ricciardi and Umezawa [36] presented a general theory of long-range coherent waves in the brain and they suggested a mechanism of memory storage and retrieval in terms of Nambu-Goldstone bosons. Jibu, Yasue and Pribram [37],[38] further developed these ideas and considered implications for consciousness. Vitiello and Freeman [39],[40] discussed memory storage in the brain in a dissipative model which is a quantum field theory model. This model is different from the Hameroff and Penrose model [41] that takes microtubules to be the place where quantum coherence is maintained.

Although the ideas of Vitiello and Freeman are more compelling than that of Hameroff and Penrose for the reason that just the existence of quantum coherence is not enough to explain subjective qualia states, they cannot be the explanation for the phenomenon of consciousness. Vitiello [39] describes the brain-environment interaction as the dialog between the brain and what he calls its "Double" as the mathematical description of the environment is obtained by the process of doubling the degrees of freedom due to entanglement. But Vitiello's dialog is one-sided and private and it cannot explain intentionality, group dynamics, and other peculiarities of sentient behavior. The fact that subjective conscious states can remain "coherent" across different individuals who are separated across space and time (as in coordinated social behavior) indicates that these states are related to a universal function. There is a dialog between this function and the neural processes inside the brain but it is not one-sided and it can embrace many individuals.

Quantum and neural processes of Figure 3 execute a dance in which the wholeness lies in the quantum part and the locus in the neural processes moves across different regions based on brain activity. The identity at the conscious level is an apparent phenomenon and the conscious self is a collection of several identities across time. This dual characteristic of the brain functioning has the potential to explain many aspects of individual and social behavior.

## DISCUSSION

We argued that quantum neural computing, a term coined to indicate a unity in the functioning of the brain, has implications for the learning modes of the brain in response to sensory input. This response may be aggregated in three types: associative, reorganizational, and quantum. We argued that the reorganizational behavior of the brain may be viewed as inner adjustment corresponding to its quantum behavior at the system level. Not only neural structures but their higher abstractions may be seen as whole entities.

We reviewed an approach to the adjustments of the weights of the interconnections that take place in a neural network. We saw the Hebbian learning mechanism as a consequence of the principle that the network studied as a whole should have one of many possible outputs.

Quantum neural computing is not about finding quantum coherent processes within the neural structure. Rather, it is the structure itself acting as a *whole* quantum system, for which the neural connections are the three-dimensional classical traces like the two-dimensional interference patterns obtained in a double-slit experiment. We are not looking for quantum characteristics of the physical cells and structures associated with neurons but rather with the quantum characteristics of the dynamics associated with the physical structures.

The dual classical/quantum conception of the brain makes it possible to see consciousness as a universal phenomenon and one can, in principle, see the plausibility of several counter-intuitive ideas associated with it. This conception has many implications: first, computers, no matter how complex, will never be conscious; second, ideas and thoughts are more than just the neural correlates of felt behavior [42].

It is customary to assume that the larger system is made up of its parts, because it is convenient to do so in the customary reductionist approach to science. In reality, one could assert with equal justification that the universe is one entity and the properties of the parts are derived from the whole.

Lecture given at *Czech Technical University, Prague* on June 25, 2009.
Arxiv paper http://arxiv.org/abs/0908.3148